\documentclass[12pt,a4paper]{article}
\usepackage[a4paper, total={7in, 10in}]{geometry}
\usepackage{iftex}
\ifPDFTeX
  \usepackage{helvet}
\else
  \usepackage{fontspec}
\fi

\usepackage{graphicx}
\usepackage[table]{xcolor}
\usepackage{float}
\usepackage{tikz}
\usetikzlibrary{arrows.meta}
\usepackage{xurl}
\usepackage{hyperref}
\hypersetup{hidelinks,breaklinks=true}
\usepackage{amsmath}
\usepackage{amssymb}
\usepackage{booktabs}
\usepackage{tabularx}
\usepackage{array}
\usepackage{orcidlink}
\usepackage{placeins}
\usepackage[super,comma,sort&compress]{natbib}

\makeatletter
\renewcommand{\maketitle}{\bgroup\setlength{\parindent}{0pt}
\begin{flushleft}
  \textbf{\@title}

  \@author
\end{flushleft}\egroup}
\makeatother

\title{Trustworthy synthetic visual media: Evidence across the media lifecycle}
\date{}

\author{%
Zexi Jia\textsuperscript{1},
Zhiqiang Yuan\textsuperscript{1},
Jie Zhou\textsuperscript{1}, and
Jinchao Zhang\textsuperscript{1,*}\\[0.5em]
\small\textsuperscript{1}Tencent WeChat AI, Beijing, China. E-mail:
\href{mailto:zxjia10@gmail.com}{zxjia10@gmail.com};
\{seraphyuan, withtomzhou, dayerzhang\}@tencent.com\\
\small\textsuperscript{*}Corresponding author: Jinchao Zhang
(E-mail: \href{mailto:dayerzhang@tencent.com}{dayerzhang@tencent.com}).}

\begin{document}

\maketitle

\section*{THE BIGGER PICTURE}

Images and videos have long helped people understand what happened and how a work came into being. Generative systems complicate that role. Realistic media can now be produced and revised without leaving a stable history, so appearance no longer reveals whether a scene was captured, synthesized, or altered along the way. Trust must instead come from evidence that explains the path an asset has taken and the circumstances in which it was used. Some of this evidence can be recovered from the media, while some must be recorded during production and preserved as the asset circulates. This review brings those approaches together and asks when their claims remain meaningful after ordinary processing or deliberate manipulation. We argue that trustworthy media do not depend on one universal marker of authenticity. The evidence must suit the question at hand, reach the person making the judgment, and remain open to correction when better information emerges. The larger goal is to keep the history of media intelligible even as the media itself continues to change.

\section*{SUMMARY}

Synthetic media rarely reach viewers in the state in which they leave a model. Editing, platform processing, and republication can each alter what a reviewer can still verify. This review asks what can be established when an asset's production history is incomplete. We introduce a \textit{claim-centered evidence framework} that compares methods by the question they can answer, the evidence available to the verifier, and the conditions under which the answer remains valid. Detection can estimate synthetic origin from the file alone, but it cannot usually recover authorship or permission. Watermarks and signed provenance records preserve a stronger connection to production when participating systems maintain that connection. Rights mechanisms address a further question: whether protected material or identity was used with authorization and what response the evidence can support. Reported benchmarks show that detector performance declines with changes in generators, subject matter, and acquisition channels, while proactive signals often fail when the workflow departs from the one for which they were designed. Verification should therefore be evaluated across the full path an asset follows. A trustworthy system keeps each claim attached to the relevant part of the asset and makes uncertainty and shared dependencies visible. It also allows a decision to change when better evidence appears.

\section*{KEYWORDS}

Synthetic Media; Content Authenticity; Media Forensics; Provenance; Digital Watermarking; Creator Rights; Trustworthy Artificial Intelligence
\section*{INTRODUCTION}

A recompressed video reaches a newsroom after passing through several accounts and platforms. Part of the scene may have been captured, another part generated, and the soundtrack replaced, yet the file carries no dependable account of those changes. The newsroom faces several questions that cannot be collapsed into one label: which parts were altered, where this version came from, whether the people depicted consented to the relevant use, and what action the available evidence justifies. A realistic appearance answers none of them by itself. Human observers can perform at near-chance levels when distinguishing GAN-synthesized faces from real ones and may even judge synthetic faces as more trustworthy, underscoring that perceptual realism is not evidence of origin.\cite{nightingale2022aisynthesized}

The difficulty is not simply to classify the asset as real or fake. Different decisions require different claims: whether content was synthesized or altered, which system or workflow produced it, whether a record of that process can be trusted, and whether the people or material involved were used with authorization. These claims draw on different evidence. Forensic detectors infer traces from the received file; watermarks and fingerprints attempt to preserve a connection to production; signed provenance records describe declared events; and rights mechanisms connect technical observations to identity, ownership, or consent.\cite{marra2019gans,wang2020cnn,wen2023treering,fernandez2023stable,c2pa2026spec,somepalli2023diffusion,shan2023glaze} Cropping, recompression, editing, and republication weaken each connection differently. What matters is which evidence survives that process and which conclusion it can still support.

Previous surveys have mapped media forensics, deepfake detection, and generated-content analysis in considerable detail.\cite{verdoliva2020media,tolosana2020deepfakes,mirsky2021creation,nguyen2022deep,masood2023deepfakes,rana2022deepfake} More recent reviews examine watermarking, proactive defenses, and the security of provenance systems.\cite{li2025digitalwatermarking,cao2025secure,kumar2025watermarking,lai2025proactive,mahara2025survey} These method-centered accounts are valuable for comparing algorithms, but they do not always reveal whether two results support the same conclusion. We instead treat a bounded claim as the unit of analysis. This perspective connects technical verification to production history and authorized use while keeping those questions distinct.

Our organizing idea is simple: each method should be judged by the claim it can support. We ask what evidence a verifier can observe and trust, how that evidence changes as the asset travels, and what action it can reasonably justify. On this basis, we bring detection, watermarking, provenance, and rights mechanisms into one claim-centered framework. We also evaluate where their links to the asset break across the media lifecycle and retain the conditions behind quantitative results, so performance can be interpreted in the workflow in which it was measured.

\subsection*{Scope of this review}

We focus on synthetic images and videos, including assets that mix captured and generated material. Audio is included when it affects the interpretation of an audiovisual claim. Interactive and three-dimensional media enter the review when their changing state alters the evidence needed for a visual conclusion. We exclude text-only detection and misinformation research without a media-evidence component, as well as legal interpretation beyond what a technical record can establish. We draw on peer-reviewed research, recent preprints, and primary technical standards, using policy and product documents only to describe formal requirements or deployed capabilities. Quantitative results are compared directly only when their sources, metrics, and evaluation protocols are sufficiently aligned. Results obtained under different protocols remain separate because their tasks and operating conditions do not support a meaningful pooled estimate.

\begin{figure}[!t]
\centering
\includegraphics[width=\textwidth]{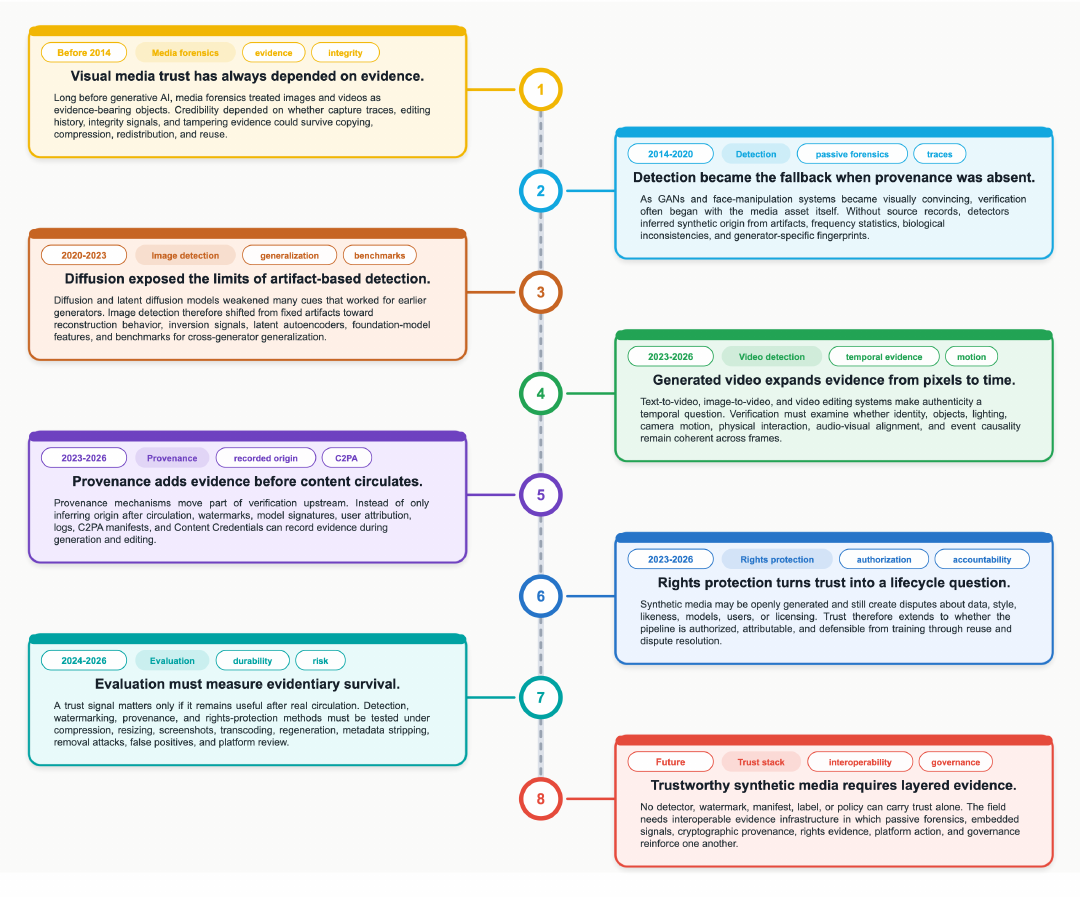}
\caption{\textbf{Evolution of evidence practices for synthetic visual media.} Milestones trace the shift from post-hoc detection to evidence recorded during creation and maintained through later use. Dates indicate when each approach became prominent rather than fixed starting points.}
\label{fig:evidence_evolution}
\end{figure}

Figure~\ref{fig:evidence_evolution} traces the field's move from artifact-based detection toward evidence that can be recorded during creation, preserved through distribution, and examined when a claim is disputed.

\FloatBarrier

The review begins by defining what counts as synthetic for a given claim and what a verifier can observe. It then moves from post-hoc detection to evidence recorded during production, preserved through distribution, and connected to rights and accountability. The final sections consider how to evaluate and combine incomplete or conflicting records, followed by a research agenda for systems that remain useful when their conclusions are challenged.

\FloatBarrier

\section*{GENERATIVE MEDIA AND THE EVIDENCE PROBLEM}

\subsection*{From generation capability to evidence demand}

Generative AI has progressed from producing isolated images to taking part in complete media workflows. Variational autoencoders and adversarial models established learned image generation, and successive advances made their outputs more stable and controllable.\cite{kingma2014autoencoding,goodfellow2014generative,radford2015unsupervised,brock2019large,karras2019style,karras2020analyzing} Diffusion and related continuous-time formulations then became a flexible basis for high-quality synthesis and editing.\cite{ho2020denoising,song2021scorebased,song2021denoising,nichol2021glide,rombach2022highresolution,podell2023sdxl,esser2024scaling} Once models could follow language and preserve user-supplied structure, generation became less a standalone task than an interface for revising visual material.\cite{radford2021learning,ramesh2021zeroshot,ramesh2022hierarchical,saharia2022photorealistic,controlnet2023AddingConditContro,instructpix2pix2023LearniFollowImage,dreambooth2023FineTuningText,openai2023dalle3,qwenimage2025technical,openai2025gpt4oimage,googledeepmind2025imagen4}

A similar shift is occurring in moving and interactive visual media. Video models now produce extended scenes in which voice, movement, and camera behavior develop together, while speech and music generators can replace the audio track that gives those scenes meaning.\cite{wavenet2016generative,dhariwal2020jukebox,audiogen2022textually,musicgen2023simple,musiclm2023generating,audioldm2023text,tulyakov2018mocogan,ho2022imagenvideo,singer2022makeavideo,blattmann2023stable,movie2024GenCastMedia,hunyuanvideo2024SystemFramewLarge,seedance2025exploring,googledeepmind2025veo} Neural rendering carried generation into three-dimensional space, while world models introduced environments that respond to action.\cite{worldmodels2018,nerf2020representing,dreamfusion2022text,pointe2022system,magic3d2022highresolution,shape2023generating,gaussian2023splatting,genie2024generative,gamengen2024diffusion,cosmos2025world,hunyuan3d2025scaling,cosmos32026omnimodal} These modalities matter here when they alter a visual claim: verification must follow not just a frame but a changing scene whose meaning may depend on sound, prior state, or user interaction.

Higher realism is only part of the change. Generation is now embedded in ordinary editing and distribution workflows, so an asset may pass through several models before it reaches a viewer. The final result may be partly captured and partly synthetic. It may also be attributable to a particular model even when its use was unauthorized. Verification must recover enough of that history to answer the specific question at stake.

\begin{figure}[!t]
\centering
\includegraphics[width=\textwidth]{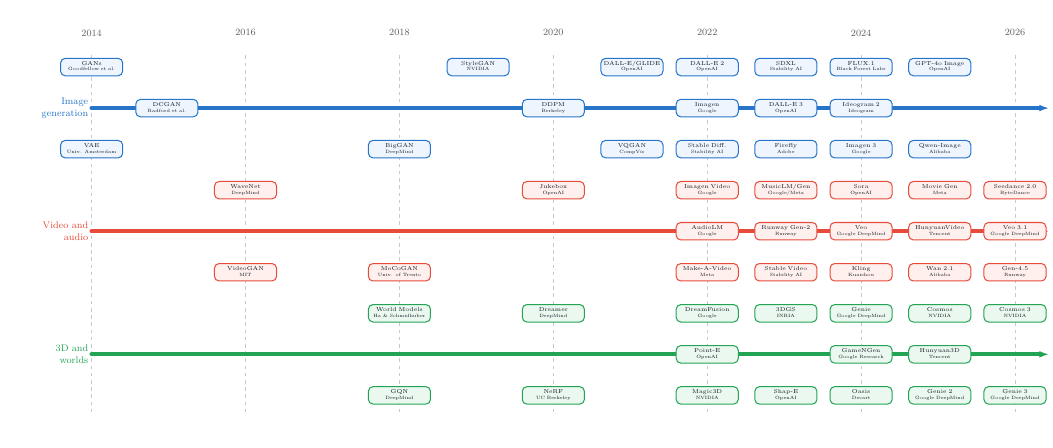}
\caption{\textbf{Expansion of generative media across visual modalities.} Selected milestones are arranged in three tracks: image generation, video and audio generation, and three-dimensional or interactive world models. Their increasing temporal, multimodal, and interactive scope broadens the conditions that verification must cover.}
\label{fig:generative_evolution}
\end{figure}

Figure~\ref{fig:generative_evolution} provides the technological backdrop. As synthesis moves beyond static pixels, verification must follow the history of an asset as its form and use evolve.

\FloatBarrier

\subsection*{Media assets and production histories}

Verification concerns two objects: the asset available for inspection and the history that produced it. We represent that distinction with compact notation:
\begin{equation}
  x=(p_x,r_x,c_x),
  \qquad
  H_x=(o_1,o_2,\ldots,o_T).
  \label{eq:asset-history}
\end{equation}
Here \(p_x\) is the observable payload, \(r_x\) contains attached or recoverable records, and \(c_x\) is the context available during review. The sequence \(H_x\) contains the operations through which the asset was produced and circulated. Two verifiers can see the same pixels yet reach different assessments when only one can validate the history behind them.

Whether an asset should be called synthetic depends on the question being asked. For a task \(\tau\), we treat \(x\) as synthetic only when a generative operation in \(H_x\) materially changes the property that \(\tau\) evaluates. This includes fully generated media as well as works altered only in the part relevant to the claim. An edit may change the answer to one claim while leaving another, such as the identity of the depicted person, unchanged.

\begin{figure}[p]
\centering
\input{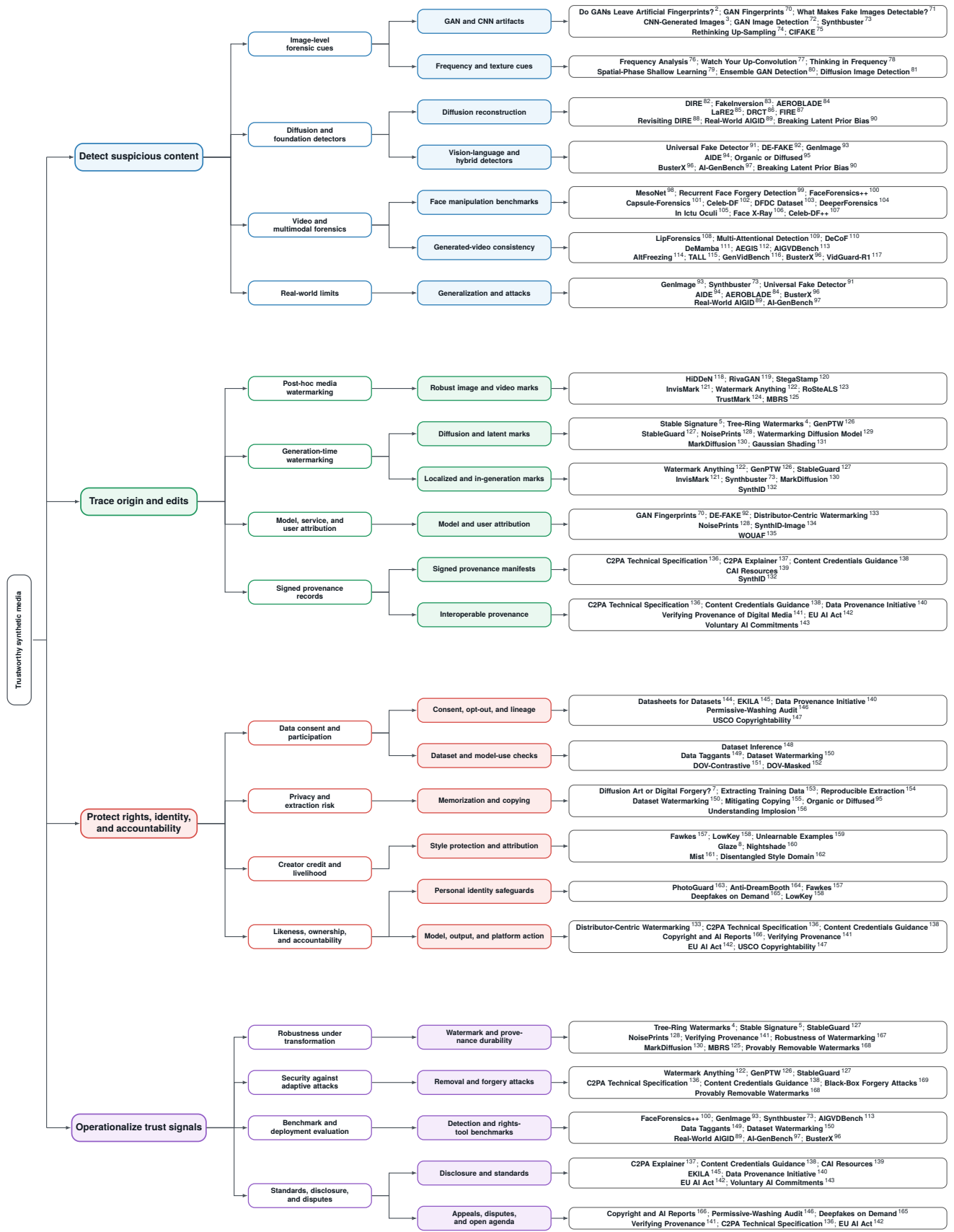}
\caption{\textbf{Claim-centered roadmap of the evidence literature.} Branches organize representative studies by lifecycle role: detection, origin and edit tracing, rights and identity protection, and operational evaluation. Within each branch, verification tasks are linked to relevant method families.}
\label{fig:field_roadmap}
\end{figure}

Figure~\ref{fig:field_roadmap} organizes the literature by the point at which evidence enters this production history and by the claim it can later support.

\nocite{marra2019gans,yu2019attributing,chai2020what,wang2020cnn,gragnaniello2021gan,bammey2024synthbuster,tan2024rethinking,bird2024cifake,frank2020leveraging,durall2020watch,qian2020thinking,liu2021spatialphase,mandelli2022detecting,corvi2023detection,wang2023dire,cazenavette2024fakeinversion,ricker2024aeroblade,luo2024lare,chen2024drct,chu2025fire,lin2025revisiting,li2025realworldaigid,genimagepp2025latent,ojha2023towards,sha2023de,zhu2023genimage,yan2025sanity,ha2024organic,busterx2025MultimLargeLangua,aigenbench2025ongoing,afchar2018mesonet,sabir2019recurrent,rossler2019faceforensics,nguyen2019capsule,li2020celebdf,dolhansky2020dfdc,jiang2020deeperforensics,li2018ictu,li2020facexray,celeb2025DfLargeScale,haliassos2021lipforensics,zhao2021multattentional,ma2024decof,chen2024demamba,li2025aegis,aigvdbench2025,wang2023altfreezing,xu2023tall,genvidbench2025,vidguard2025,zhu2018hidden,weng2019rivagan,tancik2020stegastamp,xu2025invismark,watermark2024AnythiLocaliMessag,bui2023rosteals,bui2023trustmark,jia2021mbrs,fernandez2023stable,wen2023treering,gan2025genptw,yang2025stableguard,goren2025noiseprints,zhao2023watermark,pan2025markdiffusion,yang2024gaussianshading,google2023synthid,fei2025distributor,gowal2025synthidimage,wouaf2024WeightModulaUser,c2pa2025spec,c2pa2025explainer,contentcredentials2025guidance,cai2024contentauthenticity,longpre2023dataprovenance,golaszewski2026provenance,eu2024aiact,whitehouse2023commitments,gebru2021datasheets,balan2023ekila,jewitt2026permissivewashing,usco2025part2,dziedzic2022datasetinference,bouaziz2025datataggants,wang2025datasetwatermarking,xie2025dovcontrastive,xie2025dovmasked,somepalli2023diffusion,carlini2023extracting,webster2023reproducible,somepalli2023understanding,ding2024implosion,shan2020fawkes,lowkey2021LeveraAdversAttack,huang2021unlearnable,shan2023glaze,shan2024nightshade,mist2023TowardImprovAdvers,huang2024disentangled,salman2023photoguard,vanle2023antidreambooth,hawkins2025deepfakesondemand,usco2025part3,robustness2024WatermTextImage,zhao2024removable,black2025BoxForgerAttack}

\subsection*{A claim-centered evidence framework}

Our \textit{claim-centered evidence framework} begins with a bounded proposition about an asset and asks how the available signals change confidence in it. Keeping that proposition explicit places technical measurements and institutional judgments in the same analytical structure while preserving the distinction between them.

\begin{figure}[!t]
\centering
\includegraphics[width=\textwidth]{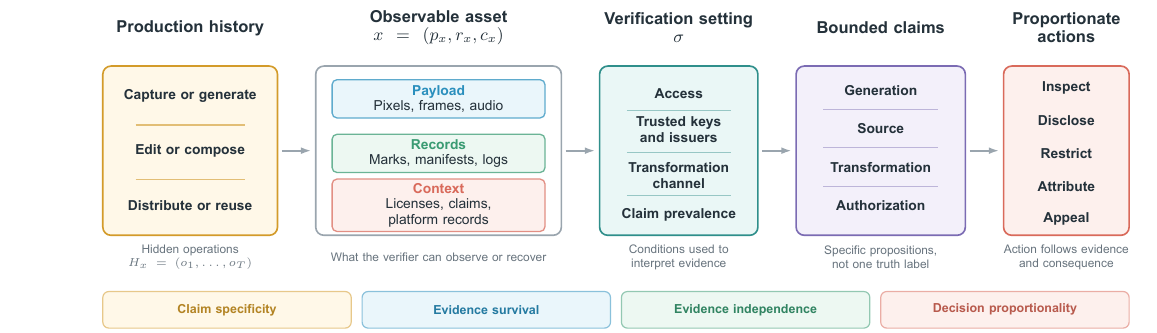}
\caption{\textbf{Claim-centered evidence framework.} Production history and observable evidence are interpreted within a verification setting to support bounded claims and proportionate actions. The lower band summarizes the four principles used to evaluate this process.}
\label{fig:claim_evidence_model}
\end{figure}

Let \(q\) be one bounded claim about the hidden history of the asset, such as synthetic origin, source attribution, or authorized use. Under verification setting \(\sigma\), a verifier observes a set of signals \(S(x,\sigma)\) from the payload, records, and context. We represent the verification process as
\begin{equation}
  x
  \ \longrightarrow\ 
  S(x,\sigma)
  \ \longrightarrow\ 
  E_v(q\mid x,\sigma)
  \ \longrightarrow\ 
  a.
  \label{eq:evidence-chain}
\end{equation}
Each transition requires an explicit inference; none is automatic. Here, \(E_v\) is the verifier's assessment of claim \(q\), not a universal trust score. The action \(a\) depends on both the uncertainty in that assessment and the consequence of an error. A detector score may be sufficient to prioritize review yet far too weak for a public accusation. Likewise, a valid manifest can support chain-of-custody without establishing that the depicted event occurred. Figure~\ref{fig:claim_evidence_model} expands this short chain.

\subsection*{Evidence roles across the lifecycle}

Four activities shape evidence across the lifecycle, each at a different moment. Detection works backward from observed media when little history survives. Provenance records part of that history before it is lost. Rights and accountability processes determine how the history bears on an affected party, and operational evaluation tests whether the evidence remains useful after distribution. None substitutes for the others: an asset can have a known source and uncertain permission, or a valid signed history that omits the event under dispute.

A layered system may combine signals from detection, provenance, rights records, and operational tests in \(S(x,\sigma)\). Agreement is informative only when the dependencies among those signals are understood. Two apparently independent results may fail together after regeneration, while an authenticated record may conflict with another source or omit a relevant event. Any combined assessment should keep those dependencies and disagreements explicit instead of hiding them in one score.

\subsection*{Evidence availability and survival}

The verification setting \(\sigma\) summarizes four practical conditions: what the verifier can access, which issuers or systems it trusts, the path through which the asset has passed, and how prevalent the claim is expected to be. At one extreme, the verifier sees only pixels. Participating systems may expose progressively richer records, whereas an adversarial setting also allows those records to be removed or forged.

Evidence must also survive the media channel. Let \(\Pi_T\) describe the transformations likely along the relevant distribution path, and let \(V_s\) test whether signal \(s\) can still be recovered. Its durability is
\begin{equation}
  D(s\mid\sigma)
  = \Pr_{T\sim\Pi_T}
  \left[V_s(T(x))=1 \mid V_s(x)=1\right].
  \label{eq:durability}
\end{equation}
A durability of 0.9 means that the signal is expected to remain recoverable after 90\% of transformations drawn from the relevant distribution path, conditional on being detectable beforehand. This measure concerns the route through which media actually travels, not merely a fixed distortion suite. Survival is still insufficient when the signal lacks a trusted issuer, supports an ambiguous claim, or produces too many false positives for the expected prevalence of the claim.

\subsection*{Principles for trustworthy evidence}

The framework yields four practical principles. \textbf{Claim specificity} keeps origin, factual truth, attribution, and authorization separate. \textbf{Evidence survival} asks whether a signal remains useful after realistic transformation and handoff. \textbf{Evidence independence} makes shared assumptions and failure modes visible when several signals are combined. \textbf{Decision proportionality} matches the strength and reviewability of the evidence to the consequence of acting on it.

These three equations provide only the notation needed for the argument. Equation~\eqref{eq:asset-history} separates the observable asset from its hidden history, Equation~\eqref{eq:evidence-chain} follows the path from signals to a bounded decision, and Equation~\eqref{eq:durability} asks whether those signals survive the path the media actually takes. Two later equations apply the same logic to deployment problems where the numerical relationship is itself informative.

\FloatBarrier

\section*{DETECTING SYNTHETIC CONTENT FROM MEDIA}

Detection is most useful when an image or video arrives without a trustworthy record of origin. It estimates whether the received content resembles the forms of synthesis or manipulation represented in its calibration data. In this setting, \(S(x,\sigma)\) consists mainly of measurements from \(p_x\), while the production history remains hidden. The result may justify closer inspection or triage, but it cannot by itself establish source, authorship, permission, or factual truth.

\subsection*{Artifacts, reconstruction, and learned cues}

Early detectors looked for regularities introduced by image synthesis, especially traces of upsampling and characteristic frequency behavior.\cite{marra2019gans,yu2019attributing,wang2020cnn,frank2020leveraging,durall2020watch} These cues made generated-image detection possible, yet they often reflected the implementation that produced the image. A detector could recognize one pipeline more reliably than synthetic origin itself.

Diffusion models shifted attention from fixed artifacts to reconstruction and inversion. An image that lies close to a learned manifold may be reconstructed differently from a camera image, and several detectors exploit that difference.\cite{wang2023dire,ricker2024aeroblade,chen2024drct,luo2024lare,cazenavette2024fakeinversion} Reconstruction behavior still depends on the model and on how the asset was edited, so the cue can change with the production pipeline.

Recent detectors combine local forensic evidence with broader visual representations, often learned from natural images or vision--language supervision.\cite{liu2022lnp,ojha2023towards,liu2024fatformer,yan2025sanity,zhou2025aigiholmes} This improves transfer across some generator families, but it can replace one shortcut with another when the training data makes content or style predictive of the label. No cue generalizes merely because it operates at a high level; its value depends on the shifts it has survived.

\subsection*{Generalization is the central detection problem}

Let \(e\) denote an environment defined by a generator, content domain, transformation channel, and acquisition process. A deployment-oriented detector should be judged by its weakest relevant environment, not only its average result:
\begin{equation}
  R_{\mathrm{rob}}(f)
  =\sup_{e\in\mathcal{E}_{\mathrm{test}}}R_e(f),
  \qquad
  R_e(f)=\mathbb{E}_{(x,y)\sim P_e}\!\left[\ell(f(x),y)\right].
  \label{eq:robust-risk}
\end{equation}
Average accuracy over familiar generators says little about the worst environment. Moving to a new generator tests dependence on the synthesis process, whereas a new domain tests whether content has become a shortcut. Re-digitization changes the signal again by introducing a new acquisition channel. The word ``generalization'' is useful only when the shift being measured is stated.

Recent benchmarks widen the range of generators and acquisition conditions used for this test.\cite{zhu2023genimage,bammey2024synthbuster,yan2025sanity,boychev2024imaginet,hong2025wildfake,genimagepp2025latent,jia2026coda,li2025realworldaigid} NTIRE 2026, for example, combines outputs from 42 generators with 36 transformations, while the SAFE Image Authenticity Challenge asks systems to detect, classify, and localize both partial and fully synthetic content.\cite{gushchin2026ntire,nguyen2026safe} Text-rich benchmarks expose another blind spot: performance varies sharply across layouts and can deteriorate after JPEG compression.\cite{wang2026textrich}

Method development is moving in the same direction. Layer-transition consistency, editing fingerprints, and quality-aware aggregation each address a different source of failure: unseen generators, post-production history, or repeated online reposting.\cite{yang2026layerconsistency,wu2026editprint,guillaro2026quality} Training on thousands of community-released generators adds further diversity, while out-of-the-box studies reveal how public detectors behave without benchmark-specific retraining.\cite{park2025community,ren2026outofbox} Together, these studies expose a basic problem: a detector may learn file handling or dataset construction instead of synthesis. Benchmarks must expose those nuisance variables and control them where possible.

Table~\ref{tab:image_detection_methods} organizes representative detectors by their evidence cue and backbone. Its three blocks separate conventional benchmark transfer, unseen-model and cross-domain transfer, and real-world re-digitization so that each numerical comparison retains a shared protocol.

\FloatBarrier
\begin{table}[!t]
\centering
\caption{\textbf{Image detection across shared evaluation protocols.} The three blocks examine cross-generator transfer, model and domain transfer, and performance after sharing or re-digitization. ForenSynths, Ojha, GenImage, and FakeForm report accuracy and average precision (Acc./AP, \%); RRDataset reports accuracy (\%).}
\label{tab:image_detection_methods}
\fontsize{6.8}{8.0}\selectfont
\setlength{\tabcolsep}{1.4pt}
\renewcommand{\arraystretch}{1.1}
\begin{tabularx}{\textwidth}{
  >{\raggedright\arraybackslash}p{0.145\textwidth}
  >{\raggedright\arraybackslash}X
  >{\raggedright\arraybackslash}p{0.14\textwidth}
  *{3}{>{\centering\arraybackslash}p{0.12\textwidth}}}
\toprule
\rowcolor{black!10}
\multicolumn{6}{l}{\textbf{Benchmark transfer: mean Acc./AP across each test set}} \\
\rowcolor{black!3}
\textbf{Method} & \textbf{Evidence cue} & \textbf{Backbone} & \textbf{ForenSynths} & \textbf{Ojha} & \textbf{GenImage} \\
\midrule
\textbf{CNNSpot} (2020)\cite{wang2020cnn} & augmentation artifacts & ResNet-50 & 76.0/86.7 & 52.8/67.5 & 53.3/63.6 \\
\textbf{FreDect} (2020)\cite{frank2020leveraging} & Fourier spectrum & ResNet-50 & 78.9/76.6 & 54.5/49.6 & 41.2/46.7 \\
\textbf{LGrad} (2023)\cite{tan2023learning} & image gradients & ResNet-50 & 86.1/91.6 & 90.9/97.3 & 61.8/62.0 \\
\textbf{UnivFD} (2023)\cite{ojha2023towards} & CLIP semantics & CLIP ViT-L/14 & 89.1/98.2 & 86.7/94.3 & 70.6/83.7 \\
\textbf{PatchCraft} (2023)\cite{zhong2023patchcraft} & texture-patch contrast & patch CNN & 84.8/92.7 & 84.0/93.8 & 86.3/96.4 \\
\textbf{FreqNet} (2024)\cite{tan2024freqnet} & learned frequency cues & ResNet-50 & 91.5/97.9 & 89.6/94.9 & 74.1/82.6 \\
\textbf{NPR} (2024)\cite{tan2024rethinking} & neighbor residuals & ResNet-50 & 92.4/95.6 & 95.1/97.4 & 77.2/83.0 \\
\textbf{FatFormer} (2024)\cite{liu2024fatformer} & forgery-aware features & CLIP + transformer & 98.2/99.5 & 93.6/98.4 & 76.7/86.7 \\
\textbf{SAFE} (2025)\cite{li2025safe} & transformation stability & image encoder & 96.0/98.7 & 95.7/99.0 & 95.5/99.1 \\
\textbf{CoDA} (2026)\cite{jia2026coda} & color-response statistics & compact ResNet & 98.2/99.6 & 97.5/99.4 & 95.9/99.1 \\
\bottomrule
\end{tabularx}

\vspace{2pt}
\begin{tabularx}{\textwidth}{
  >{\raggedright\arraybackslash}p{0.145\textwidth}
  >{\raggedright\arraybackslash}X
  >{\raggedright\arraybackslash}p{0.14\textwidth}
  *{2}{>{\centering\arraybackslash}p{0.15\textwidth}}}
\rowcolor{black!10}
\multicolumn{5}{l}{\textbf{FakeForm transfer: mean Acc./AP across 13 unseen models and 62 visual domains}} \\
\rowcolor{black!3}
\textbf{Method} & \textbf{Evidence cue} & \textbf{Backbone} & \textbf{Model} & \textbf{Domain} \\
\midrule
\textbf{CNNSpot} (2020)\cite{wang2020cnn} & augmentation artifacts & ResNet-50 & 66.8/79.3 & 50.4/52.3 \\
\textbf{UnivFD} (2023)\cite{ojha2023towards} & CLIP semantics & CLIP ViT-L/14 & 77.2/85.8 & 71.2/83.8 \\
\textbf{FreqNet} (2024)\cite{tan2024freqnet} & learned frequency cues & ResNet-50 & 78.2/88.4 & 61.3/66.3 \\
\textbf{NPR} (2024)\cite{tan2024rethinking} & neighbor residuals & ResNet-50 & 74.8/81.1 & 68.4/82.1 \\
\textbf{FatFormer} (2024)\cite{liu2024fatformer} & forgery-aware features & CLIP + transformer & 83.1/91.7 & 76.5/87.2 \\
\textbf{SAFE} (2025)\cite{li2025safe} & transformation stability & image encoder & 88.4/92.3 & 58.6/67.1 \\
\textbf{AIDE} (2025)\cite{yan2025sanity} & low/high-level fusion & CLIP + CNN & 90.8/92.6 & 67.0/83.0 \\
\textbf{AIGI-Holmes} (2025)\cite{zhou2025aigiholmes} & artifacts + reasoning & multimodal LLM & 91.8/92.8 & 73.6/86.4 \\
\textbf{CoDA} (2026)\cite{jia2026coda} & color-response statistics & compact ResNet & 91.0/93.0 & 77.7/88.1 \\
\bottomrule
\end{tabularx}

\vspace{2pt}
\begin{tabularx}{\textwidth}{
  >{\raggedright\arraybackslash}p{0.145\textwidth}
  >{\raggedright\arraybackslash}X
  >{\raggedright\arraybackslash}p{0.14\textwidth}
  *{2}{>{\centering\arraybackslash}p{0.15\textwidth}}}
\rowcolor{black!10}
\multicolumn{5}{l}{\textbf{RRDataset: accuracy under the original and re-digitized channels (\%)}} \\
\rowcolor{black!3}
\textbf{Method} & \textbf{Evidence cue} & \textbf{Backbone} & \textbf{Overall} & \textbf{Re-digitized} \\
\midrule
\textbf{CNNSpot} (2020)\cite{wang2020cnn} & augmentation artifacts & ResNet-50 & 74.3 & 43.1 \\
\textbf{GramNet} (2020)\cite{liu2020gramnet} & texture statistics & Gram-CNN & 75.4 & 79.5 \\
\textbf{FreDect} (2020)\cite{frank2020leveraging} & Fourier spectrum & ResNet-50 & 68.7 & 46.3 \\
\textbf{Fusing} (2022)\cite{ju2022fusing} & global--local artifacts & dual CNN & 65.2 & 30.8 \\
\textbf{LGrad} (2023)\cite{tan2023learning} & image gradients & ResNet-50 & 57.8 & 14.7 \\
\textbf{DNF} (2023)\cite{zhang2023dnf} & inverse-diffusion noise & ResNet-50 & 79.2 & 0.1 \\
\textbf{DIRE} (2023)\cite{wang2023dire} & reconstruction error & ResNet-50 & 79.4 & 1.4 \\
\textbf{UnivFD} (2023)\cite{ojha2023towards} & CLIP semantics & CLIP ViT-L/14 & 59.5 & 36.2 \\
\textbf{NPR} (2024)\cite{tan2024rethinking} & neighbor residuals & ResNet-50 & 67.0 & 38.7 \\
\textbf{SSP} (2024)\cite{chen2024ssp} & patch noise & compact CNN & 58.2 & 32.6 \\
\textbf{FreqNet} (2024)\cite{tan2024freqnet} & learned frequency cues & ResNet-50 & 62.8 & 37.5 \\
\textbf{DRCT} (2024)\cite{chen2024drct} & reconstruction contrast & ConvNeXt-B & 89.6 & 64.3 \\
\textbf{C2P-CLIP} (2025)\cite{tan2025c2pclip} & category prompts & CLIP ViT-L/14 & 58.6 & 18.0 \\
\textbf{SAFE} (2025)\cite{li2025safe} & transformation stability & image encoder & 64.6 & 2.3 \\
\textbf{AIDE} (2025)\cite{yan2025sanity} & low/high-level fusion & CLIP + CNN & 78.4 & 76.0 \\
\bottomrule
\end{tabularx}
\end{table}

\FloatBarrier

The three blocks expose distinct and increasingly demanding forms of transfer. In the first, FatFormer falls from 98.2\% accuracy on ForenSynths to 76.7\% on GenImage, while GenImage accuracy across the compared methods spans 41.2--95.9\%. The FakeForm block then compares unseen-model and cross-domain transfer under one evaluation protocol. The direction and size of the gap vary by detector: SAFE moves from 88.4\% accuracy across unseen models to 58.6\% across domains, and AIDE from 90.8\% to 67.0\%. The resulting gaps show that transfer across generators and transfer across visual domains are distinct properties. Re-acquisition can be still more disruptive. DRCT reaches 89.6\% overall after adaptation to RRDataset, yet established frequency and reconstruction methods fall to 0.1--46.3\% on re-digitized fakes. Accuracy and AP also diverge in several cells, showing that a useful ranking can coexist with a decision threshold that no longer transfers.

Within the framework, these values support a synthetic-origin claim only for the verification settings tested; they do not recover a source or an operation history. The sharpest drops occur when the setting changes what reaches the detector. For deployment, calibration and abstention matter more than the highest clean-test score.

\subsection*{Temporal evidence in video}

Video forensics grew out of face manipulation. Early methods tracked blinking, head motion, or localized blending, and successive benchmarks made those tests harder by varying identities, compression, and recording conditions.\cite{guera2018deepfake,rossler2019faceforensics,dang2020detection,li2020celebdf,dolhansky2020dfdc,jiang2020deeperforensics,haliassos2022leveraging} Full-scene generators change the task. They synthesize the setting, camera motion, and action around a subject, so a face-specific cue may cover only a small part of the claim.

Current detectors draw on three temporal scales. Frame-level methods aggregate spatial traces that recur across a clip. Sequence models look for motion or appearance that changes implausibly between frames, while multimodal methods compare the visible action with speech, sound, or a textual account.\cite{bohacek2024lostintranslation,ma2024decof,chen2024demamba,li2025aegis,aigvdbench2025} These signals are complementary rather than interchangeable. A frame artifact may identify a generator family without locating an altered interval, whereas a temporal inconsistency may reveal that the clip is unstable without identifying its source.

The reported results show how strongly the answer depends on the test construction. On the GVF benchmark, image detectors used without retraining achieved 35.0--53.6\% mean accuracy across four generator subsets; retraining on one text-to-video source raised the strongest baselines only to 63.3--66.7\%. DeCoF instead reported 85.9--96.4\% mean accuracy across Gen-2, Pika, Sora, Veo, and Kling, depending on which open-source generator supplied the training data.\cite{ma2024decof} The larger AEGIS study presents a different picture. On its hard set, Qwen2.5-VL variants detected only 22--23\% of synthetic videos without task-specific training; fine-tuning raised the 7B model's macro-F1 from 0.43 to 0.82 in-domain but only from 0.52 to 0.55 on the hard split.\cite{li2025aegis} AIGVDBench broadens the comparison to 31 generators, more than 440,000 videos, and 33 detectors. Its task-specific results also change the ranking: I3D reaches 89.1\% AUC on image-to-video outputs but 61.2\% on closed-source generators, while TimeSformer moves from 75.5\% to 86.5\%.\cite{aigvdbench2025}

These numbers do not establish a stable ordering of video detectors. Duration, generation task, frame sampling, the choice of real videos, and whether the generator is open or proprietary all change what reaches the classifier. Video evaluation needs interval-level ground truth and a video-level decision rule, with separate results for face manipulation, partial editing, and fully generated scenes. Source attribution is a further claim: recent work can distinguish generation task, model family, and precise generator, but that conclusion remains tied to the candidate sources represented during training.\cite{kundu2026saga}

\subsection*{What passive detection can establish}

Operational reliability depends on prevalence as well as benchmark accuracy. If synthetic media occur with prevalence \(\pi\), a detector with true-positive rate \(\mathrm{TPR}\) and false-positive rate \(\mathrm{FPR}\) has positive predictive value
\begin{equation}
  \mathrm{PPV}=\frac{\pi\,\mathrm{TPR}}
  {\pi\,\mathrm{TPR}+(1-\pi)\,\mathrm{FPR}}.
  \label{eq:ppv}
\end{equation}
For example, with 1\% prevalence, 90\% TPR, and 1\% FPR, only about 48\% of flagged items are expected to be synthetic. When \(\pi\) is small, even an apparently accurate detector can generate more false allegations than correct detections. Balanced-test accuracy is a poor deployment guide. The relevant question is whether the score remains calibrated at a sufficiently low false-positive rate and whether uncertain cases can be withheld for review.

Passive detection provides a probability about an observed asset under stated conditions. Because failure to find evidence of manipulation is not evidence that an asset is authentic, a negative result leaves open the possibility of an unseen generator.\cite{farid2025mitigating} A positive result, meanwhile, says little about source or intent. Detection is best used to trigger closer review when no stronger record survives.

\FloatBarrier

\section*{PRESERVING ORIGIN AND TRANSFORMATION HISTORY}

Provenance preserves information that would otherwise have to be inferred after the fact. It can document origin, transformation, or continuity between an observed asset and a production record. In terms of the framework, it adds records to \(S(x,\sigma)\), but only when the verifier can reach and trust the relevant keys, issuers, or remote services. A surviving watermark may justify attribution to a generator or service without explaining what happened later, while a signed history can be authentic and still omit an event that matters. Provenance supports relationships that were recorded; it cannot rule out events that were not.

\subsection*{Post-hoc watermarks and the robustness tradeoff}

Post-hoc watermarking inserts a machine-readable signal after creation, so content can be marked without changing the generator. Existing methods make different compromises between how much they encode, how visible the change is, and how well the signal survives.\cite{zhu2018hidden,weng2019rivagan,tancik2020stegastamp,bui2023trustmark,lu2025vine,xu2025invismark} Because the mark is added separately, its evidentiary value depends on what remains after the asset has passed through the rest of the workflow.

No watermark maximizes quality, capacity, and durability at once. A larger message may be easier to disrupt, while aggressive robustness training can leave a more visible trace. More importantly, surviving compression does not imply survival after the image has been regenerated in a way that preserves its semantics. Evaluation should follow the workflow the mark is expected to encounter and determine whether the recovered message still leads to an identifiable issuer and a clear claim.

\subsection*{Watermarks integrated into generation}

Generation-integrated methods use the synthesis process itself to carry evidence. Early work showed that a diffusion model could introduce a persistent pattern through its noise, decoder, or latent trajectory.\cite{recipe2023WatermDiffusModels,wen2023treering,fernandez2023stable,yang2024gaussianshading,zhang2024zodiac} Later systems moved beyond simple detection. Some localize altered regions or recover richer messages, while others are designed explicitly against removal and spoofing or adapt the idea to autoregressive image generators.\cite{gan2025genptw,yang2025stableguard,luo2026grow,yang2026pai,lukovnikov2026clustermark}

Recent work makes the security question more explicit. MaxMark increases message capacity within the diffusion latent, whereas ISTS changes both embedding and verification to address removal and forgery together.\cite{chang2026maxmark,zhu2026ists} These advances do not remove the need for adversarial evaluation. A recent audit of autoregressive watermarks shows that one marked reference image can be enough to enable removal or mimicry without access to model parameters or secret keys.\cite{muller2026arwatermark}

Embedding during generation can mark every output of a participating service without a separate encoding step. Its coverage, however, is a property of deployment rather than of the decoder alone. A model can be run without the marking component, and different services may not recognize one another's keys. Generation-time watermarking can provide strong positive evidence inside a participating ecosystem, but an absent mark remains inconclusive outside it.

Table~\ref{tab:watermark_methods} separates results obtained under shared evaluations from values retained from original protocols. This distinction prevents robustness under compression, semantic editing, and regeneration from being treated as the same property.

\FloatBarrier
\begin{table}[!t]
\centering
\caption{\textbf{Image watermarking across shared and original evaluation protocols.} W-Bench uses TPR at 0.1\% FPR, and UltraEdit uses bit accuracy; other blocks use the quality measures and test conditions of the original studies. WDP denotes watermark detection probability, and PSNR/SSIM denotes peak signal-to-noise ratio and structural similarity.}
\label{tab:watermark_methods}
\fontsize{7.6}{9.0}\selectfont
\setlength{\tabcolsep}{2.2pt}
\renewcommand{\arraystretch}{1.1}
\begin{tabularx}{\textwidth}{
  >{\raggedright\arraybackslash}p{0.20\textwidth}
  >{\centering\arraybackslash}p{0.09\textwidth}
  >{\centering\arraybackslash}p{0.12\textwidth}
  *{4}{>{\centering\arraybackslash}X}}
\toprule
\rowcolor{black!10}
\multicolumn{7}{l}{\textbf{Post-hoc methods: W-Bench}} \\
\rowcolor{black!3}
\textbf{Method} & \textbf{Payload} & \textbf{PSNR/SSIM} &
\textbf{Regeneration} & \textbf{Global edit} & \textbf{Local edit} & \textbf{Video} \\
\midrule
\textbf{RivaGAN} (2019)\cite{weng2019rivagan} & 32 bits & 40.43/0.970 & 11.3 & 14.8 & 45.6 & 3.2 \\
\textbf{StegaStamp} (2020)\cite{tancik2020stegastamp} & 100 bits & 29.65/0.911 & 91.6 & 78.7 & 99.0 & 30.9 \\
\textbf{MBRS} (2021)\cite{jia2021mbrs} & 30 bits & 27.37/0.894 & 99.4 & 59.8 & 94.4 & 13.6 \\
\textbf{CIN} (2022)\cite{ma2022cin} & 30 bits & 43.19/0.985 & 48.3 & 45.6 & 58.7 & 2.9 \\
\textbf{PIMoG} (2022)\cite{fang2022pimog} & 30 bits & 37.72/0.986 & 77.0 & 64.9 & 69.3 & 14.3 \\
\textbf{SepMark} (2023)\cite{wu2023sepmark} & 30 bits & 35.48/0.981 & 67.5 & 74.1 & 95.0 & 8.8 \\
\textbf{TrustMark} (2023)\cite{bui2023trustmark} & 100 bits & 41.27/0.991 & 21.7 & 69.0 & 68.2 & 39.6 \\
\textbf{VINE-B} (2025)\cite{lu2025vine} & 100 bits & 40.51/0.995 & 95.1 & 88.8 & 94.6 & 25.4 \\
\textbf{VINE-R} (2025)\cite{lu2025vine} & 100 bits & 37.34/0.993 & 99.8 & 93.0 & 96.5 & 36.3 \\
\midrule
\rowcolor{black!10}
\multicolumn{7}{l}{\textbf{Generation-integrated methods: UltraEdit}} \\
\rowcolor{black!3}
\textbf{Method} & \textbf{Payload} & \textbf{PSNR/SSIM} &
\textbf{Prompt edit} & \textbf{Regeneration} & \textbf{Inpainting} & \textbf{LaMa} \\
\midrule
\textbf{Stable Signature} (2023)\cite{fernandez2023stable} & 48 bits & 31.43/0.834 & 0.561 & 0.626 & 0.905 & 0.894 \\
\textbf{WOUAF} (2024)\cite{wouaf2024WeightModulaUser} & 64 bits & 30.71/0.847 & 0.587 & 0.601 & 0.874 & 0.883 \\
\textbf{LaWa} (2024)\cite{rezaei2024lawa} & 48 bits & 35.14/0.821 & 0.591 & 0.629 & 0.892 & 0.897 \\
\textbf{GenPTW} (2026)\cite{gan2025genptw} & 64 bits & 39.56/0.892 & 0.969 & 0.974 & 0.990 & 0.994 \\
\bottomrule
\end{tabularx}

\vspace{3pt}
\begin{tabularx}{\textwidth}{
  >{\raggedright\arraybackslash}p{0.20\textwidth}
  >{\centering\arraybackslash}p{0.09\textwidth}
  >{\centering\arraybackslash}p{0.13\textwidth}
  >{\raggedright\arraybackslash}p{0.20\textwidth}
  >{\raggedright\arraybackslash}X}
\rowcolor{black!10}
\multicolumn{5}{l}{\textbf{Post-hoc methods: original protocols}} \\
\rowcolor{black!3}
\textbf{Method} & \textbf{Payload} & \textbf{Quality} & \textbf{Test} & \textbf{Reported result} \\
\midrule
\textbf{WAM} (2025)\cite{watermark2024AnythiLocaliMessag} & 32 bits & 46.05/1.000 & common; paraphrase & 0.96--0.98; 0.56--0.63 WDP \\
\textbf{InvisMark} (2025)\cite{xu2025invismark} & 256 bits & 51.0/0.998 & manipulation suite & $>97\%$ bit accuracy \\
\textbf{SynthID-Image} (2025)\cite{gowal2025synthidimage} & detection & deployed & Google services & $>10$ billion images/frames \\
\textbf{PECCAVI} (2026)\cite{dixit2026peccavi} & detection & 29.84/0.930 & common; paraphrase & 0.91--0.99; 0.85--0.90 WDP \\
\bottomrule
\end{tabularx}

\vspace{3pt}
\begin{tabularx}{\textwidth}{
  >{\raggedright\arraybackslash}p{0.20\textwidth}
  >{\centering\arraybackslash}p{0.09\textwidth}
  >{\centering\arraybackslash}p{0.13\textwidth}
  >{\raggedright\arraybackslash}p{0.20\textwidth}
  >{\raggedright\arraybackslash}X}
\rowcolor{black!10}
\multicolumn{5}{l}{\textbf{Generation-integrated methods: original protocols}} \\
\rowcolor{black!3}
\textbf{Method} & \textbf{Payload} & \textbf{Quality} & \textbf{Carrier} & \textbf{Reported result} \\
\midrule
\textbf{Tree-Ring} (2023)\cite{wen2023treering} & detection & 25.77/0.920 & diffusion noise & 0.92--0.98 / 0.68--0.77 \\
\textbf{Gaussian Shading} (2024)\cite{yang2024gaussianshading} & multi-bit & 30.23/0.920 & diffusion latent & 0.93--0.99 / 0.71--0.81 \\
\textbf{ZoDiac} (2024)\cite{zhang2024zodiac} & detection & 28.47/0.920 & latent optimization & 0.89--0.92 / 0.70--0.81 \\
\textbf{MaxMark} (2026)\cite{chang2026maxmark} & high capacity & matched & diffusion latent & up to 46\% bit-accuracy gain \\
\textbf{ISTS} (2026)\cite{zhu2026ists} & detection & prompt-conditioned & dynamic noise & 0.936/0.58 removal AUC/TPR \\
\textbf{GROW} (2026)\cite{luo2026grow} & 16 bits & 27.54/0.850 & diffusion sampling & 97.8\% message accuracy \\
\textbf{PAI} (2026)\cite{yang2026pai} & key & FID 29.27 & diffusion trajectory & 99.0\% removal; 96.3\% spoofing \\
\textbf{IndexMark} (2025)\cite{tong2025indexmark,lukovnikov2026clustermark} & detection & FID 4.49 & autoregressive tokens & 83.4\% regeneration TPR \\
\textbf{WMAR} (2025)\cite{jovanovic2025wmar,lukovnikov2026clustermark} & detection & FID 4.23 & tokens + tuning & 75.8\% regeneration TPR \\
\textbf{NoisePrints} (2025)\cite{goren2025noiseprints} & seed proof & distortion-free & noise + seed & third-party image/video proof \\
\textbf{ClusterMark} (2026)\cite{lukovnikov2026clustermark} & detection & FID 4.85 & clustered tokens & 96.9\% mean TPR (6 attacks) \\
\bottomrule
\end{tabularx}
\end{table}

\FloatBarrier

The shared evaluations make the trade-offs concrete. VINE-R is more robust than VINE-B but has lower PSNR, and its TPR still falls to 36.3\% after image-to-video conversion. Under visual paraphrasing, WAM reports a detection probability of 0.56--0.63, whereas PECCAVI reports 0.85--0.90 by concentrating the mark in more stable regions. Generation-integrated methods can fare better under semantic editing: GenPTW reports 0.969--0.994 bit accuracy across UltraEdit, although only under its own 64-bit protocol. ImageDetectBench likewise compares passive and watermark-based detectors under eight routine and three adversarial perturbations; its strongest watermarks are more robust when their verifier is available, but this advantage depends on participation in the marking system.\cite{guo2026passivewatermark} No reported method combines broad generator coverage, efficient verification, and durable recovery across all of the workflows represented in Table~\ref{tab:watermark_methods}. Even when a signal reconnects an asset to an issuer, source history, later edits, and authorization still depend on the records to which that signal leads.

\subsection*{Attribution at model, service, and user levels}

Attribution can begin with a trace in the output, a mark placed during generation, or a record retained by the service. Each ties the conclusion to a different point in the production chain.\cite{yu2019attributing,fei2025distributor,goren2025noiseprints} Recent retrieval-based attribution avoids fixing the candidate set during detector training and can adapt to a new source from a few examples, while video attribution has begun to separate generator, version, task, and developer at several levels.\cite{wang2026lida,kundu2026saga} A model fingerprint may identify a family of generators, whereas an authenticated service record can narrow the claim to one deployment. Neither, by itself, identifies the person responsible for publishing the asset.

Technical attribution and human responsibility often diverge because models are redistributed and accounts do not map cleanly to individual action. The person who publishes an asset may not be the person who generated it. Attribution should stop at the narrowest level justified by the evidence and state what access made that conclusion possible. Corroborating sources can strengthen the technical link, but an institution must still connect that link to conduct before assigning responsibility.

\subsection*{Signed records and evidence continuity}

Content Credentials and C2PA represent provenance as signed, structured assertions. A record typically binds an asset identifier to a declared history, an issuer, a set of assertions, and a cryptographic signature. In the notation of Equation~\eqref{eq:asset-history}, it enriches \(r_x\) with a reviewable account of part of \(H_x\). Cryptographic verification can show that the protected assertion has not changed since signing. It cannot reveal events that were never recorded or establish that the scene itself is true.\cite{c2pa2026spec,c2pa2025explainer}

No single carrier is likely to preserve a complete history. A manifest can describe an edit precisely but may disappear when the asset is screenshotted or transcoded. A watermark carries less information, yet can survive long enough to reconnect the payload to a richer remote record.\cite{contentcredentials2025guidance,cai2024contentauthenticity} The two are most useful when they reinforce that connection rather than repeat the same claim.

Detection remains useful within this design. Missing credentials leave origin unresolved, while disagreement between the payload and a signed history gives a reviewer a concrete reason to investigate. Provenance works best when its claims can be inspected and challenged instead of being reduced to a badge. A reviewable connection is especially important when the dispute concerns a person's rights.

\FloatBarrier

\section*{PROTECTING CREATORS, IDENTITY, AND ACCOUNTABILITY}

A rights dispute begins where origin evidence stops. Knowing that an image is synthetic, or even which tool produced it, says little about permission. The relevant question is how a protected resource entered the production chain and whether the actor responsible was entitled to use it. Resolving that question requires a record of the relationship, not simply a label on the output.

One output can raise more than one claim because inclusion in a dataset, influence on a model, and reproduction in an output are not the same relationship. Authorization is a separate question. The observable evidence may include protected samples, dataset or service records, model responses, and claimant documentation; different verifiers may have access to different subsets of this evidence. Evidence that establishes one link should not stand in for the others. Its immediate role is usually to support investigation, notice, or review; stronger sanctions require a reproducible connection to the particular relationship under dispute and a route for the other party to respond.

\subsection*{Rights claims and evidentiary requirements}

A rights claim should name five elements: the actor or affected party, the operation at issue, the protected resource, the pipeline stage, and the institutional context. Specifying the operation clarifies how the resource entered the pipeline. This structure prevents evidence of dataset inclusion from being mistaken for evidence of reproduction or lack of permission. Each link remains a claim in its own right.

Technical tests become more persuasive when administrative records tie the result to a known resource and point in time. Conversely, a license matters only when it applies to the operation being examined. Strong cases connect these records and preserve the history of any challenge, rather than elevating one score above the rest.

\subsection*{Training data authorization and evidence of use}

Training-data governance begins before a model is trained. A later audit is far easier when the dataset preserves where material came from and the terms under which it was collected. Web-scale aggregation often strips away that context, making authorization difficult to reconstruct after the model exists.\cite{longpre2023dataprovenance,gebru2021datasheets} An opt-out registry can express a current preference, but it cannot reveal what happened to copies already in circulation.

After training, an external party may need to show that protected data were used in training or influenced the model. Existing tests infer this connection from the model's behavior or from signals deliberately placed in protected examples.\cite{dziedzic2022datasetinference,bouaziz2025datataggants,xie2025dovcontrastive,xie2025dovmasked,wang2025datasetwatermarking,ren2026dwbench} Black-box membership inference can operate without access to model weights, but it still supports a dataset-membership claim rather than one about authorization or downstream influence.\cite{bohacek2025genaiconfessions} The reliability of these tests changes sharply when few examples are marked or when the model has subsequently been adapted. DWBench also shows that a method can appear reliable for one claimant yet produce false claims when many owners test the same model.

Inclusion does not imply that a sample materially influenced the model or will reappear in an output. Memorization studies show that diffusion models can reveal training examples under particular prompting and sampling conditions, especially when examples are repeated.\cite{somepalli2023diffusion,carlini2023extracting,webster2023reproducible} This finding characterizes model behavior, not authorization. Failure to extract an image also leaves membership unresolved, so an audit must state which relationship its result actually tests.

\subsection*{Protecting creative work, style, and likeness}

Creator-facing defenses alter protected media before a model can learn from or manipulate it. Some make the relevant identity or style harder to learn; others disrupt the behavior of a downstream model.\cite{shan2020fawkes,lowkey2021LeveraAdversAttack,huang2021unlearnable,mist2023TowardImprovAdvers,shan2023glaze,shan2024nightshade,salman2023photoguard,vanle2023antidreambooth,liu2023metacloak,wang2024simac,guo2024rid} Their appeal is that a creator can act without the provider's cooperation. The protection is nevertheless conditional on the pipeline anticipated during development.

Recent defenses cover more than one personalization method. IDGuardian targets both identity extraction and injection, allowing the same protected portrait to resist training-based and training-free personalization. DeepProtect focuses on face swapping, whereas StyleProtect updates only style-sensitive cross-attention layers during fine-tuning.\cite{xiong2026idguardian,lee2026deepprotect,tang2026styleprotect} UniDef pursues transfer across editing models, while VisiLock replaces image perturbation with a visual key that authorizes an editing model.\cite{shao2026unidef,le2026visilock} These are distinct forms of control and should not be collapsed into a single protection score.

Style and identity claims involve different harms. Style can be recognizable across a body of work without any one image being copied. Identity misuse can likewise cause harm even when the result is not perfectly deceptive. Non-consensual generation services show that the loss of control itself can matter.\cite{hawkins2025deepfakesondemand} Protection should extend beyond one model and be paired with a record that can support a later complaint or review.

Protection effectiveness changes with the surrounding pipeline. A provider can add purification or replace an encoder, and routine platform processing may weaken the perturbation even without an adaptive attack. Results from a fixed pipeline establish effectiveness only against the models and processing steps tested, not permanent protection. Identity defenses often report face-detection failure rate (FDFR) or identity similarity (ISM), whereas style defenses rely on generation-specific or human judgments; these values should not be pooled. Protection is more durable when prevention is backed by records that can support later verification and remedy.

\subsection*{Model ownership, service attribution, and user accountability}

The model itself may also be the protected resource. Fingerprints and distributor-specific signatures can reveal whether an output or behavior descends from a protected model.\cite{fei2025distributor,wouaf2024WeightModulaUser} That lineage becomes harder to recover after merging, distillation, or fine-tuning. Cert-LAS begins to address this gap by certifying model-ownership verification under bounded parameter perturbations rather than reporting only empirical resistance.\cite{qi2026certlas} A credible ownership result should state how much modification it survives and whether an independent party can reproduce it.

Service records can connect a request to an account and preserve the model version used, while a user-specific mark may carry part of that connection beyond the service. Neither identifies a person with certainty. Accounts can be shared or compromised, and the publisher of an asset may be different from whoever initiated generation. Accountability depends on documenting those handoffs rather than treating an account identifier as proof of conduct.

Persistent identifiers help investigate abuse, but they can also expose confidential activity and vulnerable users. Rights infrastructure should reveal no more than the claim requires and retain that information only as long as its purpose justifies. Attribution should be possible when needed without making ordinary creative work permanently traceable.

\subsection*{From technical signals to rights decisions}

Technical evidence can connect an output to a protected resource, but it does not decide the dispute. Legal and platform processes interpret that connection in light of consent and the circumstances of use.\cite{usco2025part2,usco2025part3,eu2024aiact} A watermark match may justify an investigation or support a notice; the outcome rests on the wider record.

A rights record should state who is making the claim, what relationship is being tested, and how the technical result was obtained. It should also include relevant permission records, known sources of error, and a procedure for the other party to respond. Table~\ref{tab:rights_protection_methods} retains unified dataset-auditing results where they are available and separates creator, identity, and ownership mechanisms evaluated under different protocols.

\FloatBarrier
\begin{table}[!t]
\centering
\caption{\textbf{Rights, identity, and ownership protection mechanisms.} In the two DWBench blocks, TPR is measured at 5\% FPR and VSR is reported as 0 or 1; WR denotes the data-marking rate in each column header. The other two blocks use the evaluation measures reported by their respective studies.\cite{ren2026dwbench}}
\label{tab:rights_protection_methods}
\fontsize{7.6}{9.0}\selectfont
\setlength{\tabcolsep}{2.8pt}
\renewcommand{\arraystretch}{1.1}
\begin{tabularx}{\textwidth}{
  >{\raggedright\arraybackslash}p{0.22\textwidth}
  >{\raggedright\arraybackslash}p{0.17\textwidth}
  *{4}{>{\centering\arraybackslash}X}}
\toprule
\rowcolor{black!10}
\multicolumn{6}{l}{\textbf{Dataset auditing: CIFAR-10 and ResNet-18 (DWBench)}} \\
\rowcolor{black!3}
\textbf{Method} & \textbf{Signal} & \textbf{TPR (WR=1\%)} & \textbf{TPR (WR=0.01\%)} &
\textbf{VSR (WR=1\%)} & \textbf{VSR (WR=0.01\%)} \\
\midrule
\textbf{Radioactive Data} (2020)\cite{sablayrolles2020radioactive} & feature tag & 23.3 & 4.7 & 1 & 0 \\
\textbf{DVBW} (2023)\cite{li2023dvbw} & backdoor test & 92.4 & 89.4 & 1 & 1 \\
\textbf{DYTMark} (2023)\cite{tang2023dytmark} & clean-label mark & 91.7 & 60.4 & 1 & 1 \\
\textbf{ImgDup} (2024)\cite{huang2024imgdup} & optimized twins & 8.1 & 20.0 & 1 & 0 \\
\midrule
\rowcolor{black!10}
\multicolumn{6}{l}{\textbf{Dataset auditing: Pok\'emon and Stable Diffusion 1.4 LoRA (DWBench)}} \\
\rowcolor{black!3}
\textbf{Method} & \textbf{Signal} & \textbf{TPR (WR=10\%)} & \textbf{TPR (WR=2\%)} &
\textbf{VSR (WR=10\%)} & \textbf{VSR (WR=2\%)} \\
\midrule
\textbf{RIW} (2023)\cite{tan2023riw} & edit-resistant mark & 70.6 & 40.6 & 0 & 0 \\
\textbf{GenWM} (2023)\cite{ma2023genwm} & generative mark & 82.7 & 12.8 & 0 & 0 \\
\textbf{DiffusionShield} (2023)\cite{cui2023diffusionshield} & multi-bit mark & 0 & 0 & 0 & 0 \\
\textbf{DiagnoB} (2024)\cite{wang2024diagnosis} & behavioral trigger & 99.1 & 35.3 & 1 & 0 \\
\textbf{EnTruth} (2024)\cite{ren2024entruth} & semantic mark & 99.3 & 12.2 & 1 & 0 \\
\textbf{DwtWM} (2024)\cite{wang2024datasetabuse} & wavelet mark & 77.8 & 38.1 & 1 & 0 \\
\textbf{AdvWM} (2024)\cite{wang2024datasetabuse} & adversarial mark & 94.7 & 19.3 & 1 & 0 \\
\textbf{Siren} (2025)\cite{li2025siren} & early-learning signal & 98.3 & 68.7 & 1 & 0 \\
\textbf{FT-Shield} (2025)\cite{cui2025ftshield} & expert verifier & 32.7 & 14.2 & 0 & 0 \\
\bottomrule
\end{tabularx}

\vspace{2pt}
\begin{tabularx}{\textwidth}{
  >{\raggedright\arraybackslash}p{0.22\textwidth}
  >{\raggedright\arraybackslash}p{0.12\textwidth}
  >{\raggedright\arraybackslash}p{0.23\textwidth}
  >{\raggedright\arraybackslash}X}
\rowcolor{black!10}
\multicolumn{4}{l}{\textbf{Creator, style, and identity protection}} \\
\rowcolor{black!3}
\textbf{Method} & \textbf{Protects} & \textbf{Strategy} & \textbf{Original evaluation} \\
\midrule
\textbf{Fawkes} (2020)\cite{shan2020fawkes} & identity & feature cloak & PubFig: $>95\%$; APIs: 100\% \\
\textbf{LowKey} (2021)\cite{lowkey2021LeveraAdversAttack} & identity & adversarial filter & Amazon: $<1\%$ recognition \\
\textbf{Unlearnable Examples} (2021)\cite{huang2021unlearnable} & training data & training perturbation & CIFAR/faces: near-random accuracy \\
\textbf{Mist} (2023)\cite{mist2023TowardImprovAdvers} & style & transferable perturbation & Cross-model transfer; survives purification \\
\textbf{Glaze} (2023)\cite{shan2023glaze} & style & style cloak & Stable Diffusion: $>92\%$; adaptive $>85\%$ \\
\textbf{PhotoGuard} (2023)\cite{salman2023photoguard} & editing & input immunization & Stable Diffusion: FID 167.6; SSIM 0.50 \\
\textbf{Anti-DreamBooth} (2023)\cite{vanle2023antidreambooth} & identity & personalization defense & VGGFace2: FDFR 0.63/0.76; ISM 0.33/0.28 \\
\textbf{MetaCloak} (2023)\cite{liu2023metacloak} & identity & meta-poisoning & Replicate: black-box transfer \\
\textbf{Nightshade} (2024)\cite{shan2024nightshade} & style & concept poisoning & SDXL: 70--80\% at 50 poisons \\
\textbf{SimAC} (2024)\cite{wang2024simac} & identity & timestep attack & CelebA: FDFR 96.9\% \\
\textbf{RID} (2024)\cite{guo2024rid} & identity & one-pass perturbation & A100: 0.12 s/image \\
\textbf{IDGuardian} (2026)\cite{xiong2026idguardian} & identity & dual-stage perturbation & VGGFace2: PSNR 32.19; SSIM 0.842 \\
\textbf{DeepProtect} (2026)\cite{lee2026deepprotect} & identity & feature and attribute defense & Five face-swap pipelines \\
\textbf{StyleProtect} (2026)\cite{tang2026styleprotect} & style & selective attention update & WikiArt: 30 artists; cross-model tests \\
\textbf{UniDef} (2026)\cite{shao2026unidef} & editing & model-agnostic perturbation & Multiple models and editing tasks \\
\bottomrule
\end{tabularx}

\vspace{2pt}
\begin{tabularx}{\textwidth}{
  >{\raggedright\arraybackslash}p{0.22\textwidth}
  >{\raggedright\arraybackslash}p{0.12\textwidth}
  >{\raggedright\arraybackslash}p{0.23\textwidth}
  >{\raggedright\arraybackslash}X}
\rowcolor{black!10}
\multicolumn{4}{l}{\textbf{Ownership verification and attack forensics}} \\
\rowcolor{black!3}
\textbf{Method} & \textbf{Claim} & \textbf{Evidence} & \textbf{Original evaluation} \\
\midrule
\textbf{Data Taggants} (2024)\cite{bouaziz2025datataggants} & dataset use & secret-key tags & ImageNet: black-box; no accuracy loss \\
\textbf{DOV4CL} (2025)\cite{xie2025dovcontrastive} & pretraining use & relation shift & 5 contrastive models: $p < 0.05$ \\
\textbf{DOV4MM} (2025)\cite{xie2025dovmasked} & pretraining use & reconstruction shift & 14 masked models: $p < 0.05$ \\
\textbf{Dataset-WM Eval.} (2025)\cite{wang2025datasetwatermarking} & diffusion data & removal benchmark & Customized Stable Diffusion: complete removal \\
\textbf{Distributor WIC} (2025)\cite{fei2025distributor} & user attribution & output fingerprint & 4 generators: hundreds of ms \\
\textbf{DWBench} (2026)\cite{ren2026dwbench} & dataset use & unified audit & 25 methods: 7/10 fail below 1\% WR \\
\textbf{PAI} (2026)\cite{yang2026pai} & output ownership & keyed forensics & 12 attacks: 98.43\% mean \\
\textbf{RecoverMark} (2026)\cite{an2026recovermark} & face ownership & localize + recover & 6 attacks: 99.9\% ownership \\
\textbf{PECCAVI} (2026)\cite{dixit2026peccavi} & image ownership & stable-region mark & Paraphrase: WDP 0.90/0.85 \\
\textbf{Cert-LAS} (2026)\cite{qi2026certlas} & model ownership & certified smoothing & Certified under parameter perturbation \\
\bottomrule
\end{tabularx}
\end{table}

\FloatBarrier

DWBench shows that high sample-level TPR at 10\% marking does not ensure dataset-level verification at 2\% participation or with multiple claimants. The other blocks address different claims: creator-facing methods test disruption, whereas ownership methods test later identification. Their results are meaningful only with explicit access assumptions, false-positive control, and independent verification.

\subsection*{From preference to remedy}

A lifecycle rights system begins by attaching a preference or license to a resource that can still be identified later. When the resource enters a dataset or production workflow, that relationship becomes part of its record. A subsequent technical test can then reconnect the disputed model or output to the earlier history, giving a platform something concrete to review. No link is sufficient by itself: a strong technical match cannot identify a rightsholder without a claimant record, and a valid license says little unless it applies to the disputed use.

Accountability requires enough of this chain to be reconstructed that a decision can be explained and both parties can respond. Before such evidence carries consequential weight, its reliability must be tested under the conditions in which it will be used.

\FloatBarrier

\section*{EVALUATING EVIDENCE IN REAL-WORLD PIPELINES}

Evaluation must match a method to a specific claim and decision context. The same detector can be adequate for triage and unsafe for public attribution, while a valid manifest can document an edit without establishing that the depicted event occurred. Clean accuracy hides both distinctions. Any evaluation must state what the verifier can access, which conditions may change the signal, and what follows from an error.

For each method, claim, and verification setting, we record an evidence profile with five parts: required access and trust; transformations and adversaries examined; metrics and operating points; known failures; and practical cost. Two methods are directly comparable only where the relevant parts of these profiles align. This is why the preceding tables separate shared evaluations from values retained under original protocols.

Table~\ref{tab:benchmark_conditions} makes this alignment explicit. Each benchmark exposes a different break in the evidence chain, so the rows describe operating conditions rather than one leaderboard.

\begin{table}[!t]
\centering
\caption{\textbf{Benchmarks and deployment conditions across the media lifecycle.} Rows specify the claim, tested condition, primary metric, and verifier input needed to judge whether results are directly comparable.}
\label{tab:benchmark_conditions}
\fontsize{7.6}{9.0}\selectfont
\setlength{\tabcolsep}{2.2pt}
\renewcommand{\arraystretch}{1.1}
\begin{tabularx}{\textwidth}{
  >{\raggedright\arraybackslash}p{0.18\textwidth}
  >{\raggedright\arraybackslash}p{0.15\textwidth}
  >{\raggedright\arraybackslash}X
  >{\raggedright\arraybackslash}p{0.17\textwidth}
  >{\raggedright\arraybackslash}p{0.16\textwidth}}
\toprule
\rowcolor{black!3}
\textbf{Benchmark} & \textbf{Claim} & \textbf{Tested condition} & \textbf{Metric} & \textbf{Verifier input} \\
\midrule
\textbf{GenImage} (2023)\cite{zhu2023genimage} & synthetic origin & unseen generators and image classes & accuracy & image \\
\textbf{FakeForm} (2026)\cite{jia2026coda} & synthetic origin & generator and domain transfer & accuracy & image \\
\textbf{RRDataset} (2025)\cite{li2025realworldaigid} & synthetic origin & sharing and re-digitization & overall and re-digitized accuracy & image \\
\textbf{NTIRE} (2026)\cite{gushchin2026ntire} & synthetic origin & 42 generators; 36 transformations & ROC AUC & image \\
\textbf{Out-of-box} (2026)\cite{ren2026outofbox} & detector selection & 12 datasets; 291 generators & mean accuracy and rank & image + detector \\
\textbf{Text-rich} (2026)\cite{wang2026textrich} & synthetic origin & six layout domains; JPEG processing & category accuracy & image \\
\textbf{AEGIS} (2025)\cite{li2025aegis} & synthetic origin & in-domain and hyper-realistic hard sets & accuracy and macro-F1 & video or key frames \\
\textbf{AIGVDBench} (2026)\cite{aigvdbench2025} & synthetic origin & 31 generators; three generation tasks & AUC and accuracy & video \\
\textbf{W-Bench} (2025)\cite{lu2025vine} & watermark survival & semantic, local, and video edits & TPR at 0.1\% FPR & image + decoder \\
\textbf{UltraEdit} (2026)\cite{gan2025genptw} & message recovery & prompt edit, regeneration, inpainting & bit accuracy & image + decoder \\
\textbf{ImageDetectBench} (2026)\cite{guo2026passivewatermark} & synthetic origin & eight routine and three adversarial attacks & effectiveness and efficiency & image or decoder \\
\textbf{DWBench} (2026)\cite{ren2026dwbench} & dataset use & low marking rate; multiple owners & TPR and verification success & model queries + data \\
\textbf{SAFE Challenge} (2026)\cite{nguyen2026safe} & authenticity and location & partial and full synthesis & detection and localization & image \\
\bottomrule
\end{tabularx}
\end{table}

Across Tables~\ref{tab:image_detection_methods}--\ref{tab:rights_protection_methods}, three sources of failure emerge. Payload evidence changes with the generator, subject, or acquisition channel. Proactive evidence can disappear when an asset leaves the participating workflow. Rights evidence becomes statistically weak when few protected samples are available or many claimants are tested. Calling all three problems ``robustness'' obscures what an evaluation needs to vary.

\subsection*{Distribution shifts that alter evidence}

``Generalization'' is often used as though every distribution shift were equivalent. A new generator tests dependence on one synthesis process; unfamiliar subject matter exposes content shortcuts; editing and re-acquisition alter the signal after production. These shifts can move in opposite directions, as shown by the model-to-domain gaps in Table~\ref{tab:image_detection_methods}.\cite{zhu2023genimage,bammey2024synthbuster,yan2025sanity,jia2026coda,li2025realworldaigid,gushchin2026ntire,wang2026textrich}

A benchmark must also prevent the dataset from answering the question for the detector. Real and generated samples need comparable content, encoding, and processing histories, with explicit controls for metadata and compression. Video adds duration, motion, frame sampling, generation task, and audio. A face swap, a generated insert, and a fully synthesized scene require separate ground truth rather than one undifferentiated fake class.\cite{ma2024decof,chen2024demamba,li2025aegis,aigvdbench2025}

Aggregate performance becomes informative only when the individual environments remain visible. Alongside the mean, an evaluation needs the worst environment and the dispersion across environments. This reveals whether a method transfers broadly or succeeds because several easy conditions offset one deployment-breaking failure.

\subsection*{Routine processing and adaptive attacks}

Ordinary media handling estimates whether evidence survives its expected route to a viewer. An adaptive attacker chooses an operation after learning how the verifier works. The same crop, recompression, or regeneration can belong to either setting; the distinction lies in why it was selected and what the attacker knows. Mixing the two produces a robustness number that describes neither deployment nor security.

Watermark security illustrates the distinction. Conventional distortion tests estimate survival during routine processing, whereas semantic regeneration can remove the signal while preserving what a viewer recognizes. Other attacks try to forge or transplant a convincing mark, creating false attribution instead of simple evasion.\cite{zhao2024removable,gan2025genptw,lu2025vine,dixit2026peccavi,zhu2026ists,muller2026arwatermark} Provenance and rights systems face the same broader threat: an attacker can target not just the signal but the records, identities, or verification procedure that give it meaning.\cite{shao2025databench}

Security evaluation needs a threat model that states the attacker's knowledge, query access, compute, and control over registries or accounts. The target is the complete verification path, not only a detector or decoder. A watermark may survive while its issuer record is replaced, and a signature may remain valid while the signed history is incomplete.\cite{nemecek2026contradictions,golaszewski2026provenance}

\subsection*{Calibration for the intended decision}

Evidence quality depends on the operating point and the prevalence of the claim. Accuracy and area under the curve can remain high while a low-prevalence deployment produces more false allegations than correct detections. Reports should include calibration at the low false-positive rates required in practice and an abstention region where the data provide too little support. Attribution faces the same problem when the true source is absent from the candidate set.

A threshold has meaning only through the action it triggers. A sensitive detector can prioritize internal review; a public label, loss of monetization, or named attribution requires successively stronger corroboration. Ownership tests add multiple-claim risk because repeated or adaptively selected claims inflate false discovery. Their protocols need an explicit claimant population and null hypothesis, and recovery of a mark remains separate from proof that the claimant holds the asserted right.

Calibration also needs to expose uneven coverage. Confidence can shift with the depicted population, language, genre, or capture device. A precise score is misleading when the calibration set contains little evidence for the case at hand.

\subsection*{Longitudinal evaluation from creation to contest}

Provenance needs a longitudinal test. The issue is not only whether a signal survives, but whether a later verifier can still reach the correct record and interpret it after the asset has been transformed, a key revoked, or an issuer's status changed. C2PA supplies a structure for signed assertions, yet interoperability and incomplete histories remain empirical questions.\cite{c2pa2026spec,c2pa2025explainer,contentcredentials2025guidance,golaszewski2026provenance}

Longitudinal evaluation must also account for how people and institutions interpret the evidence. A provenance label can draw attention to source history while still encouraging viewers to mistake a credible record for proof that the content is true. Wording, placement, and the treatment of missing credentials shape that interpretation.\cite{gamage2025labeling,feng2023provenance,trattner2026c2palabels} Cost and access matter as well: a test that requires private weights or thousands of paid queries may be reproducible in a provider's laboratory but unavailable to a newsroom or independent creator.

\begin{figure}[!t]
\centering
\includegraphics[width=\textwidth]{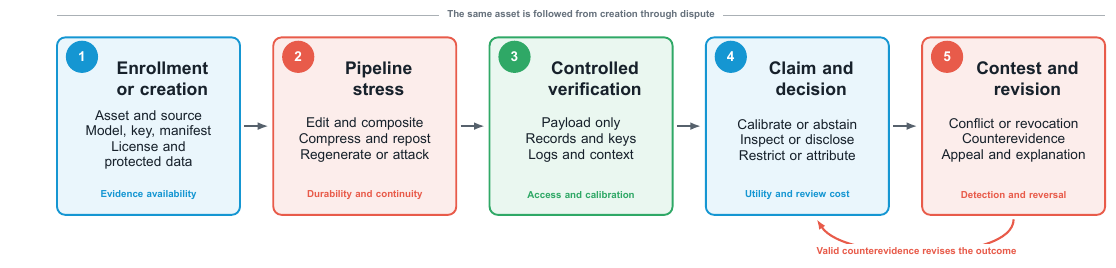}
\caption{\textbf{Lifecycle evaluation from creation to contest.} Evidence is disclosed stage by stage as an asset passes through the workflow, allowing the final decision to be tested against valid counterevidence.}
\label{fig:lifecycle_evaluation}
\end{figure}

A deployment benchmark can bring these requirements together by following one asset from creation to a contested decision. It records the true operation history, sends the asset through a realistic workflow, and reveals only the evidence that each reviewer would normally receive. The final stage introduces a missing record, conflict, revocation, or valid counterevidence and tests whether the decision updates consistently with the known history and decision rule. Figure~\ref{fig:lifecycle_evaluation} summarizes this protocol.

\FloatBarrier

Such a benchmark moves the endpoint from signal recovery to decision quality. It measures whether the claim remains calibrated, whether its supporting record remains reachable, how much review costs, and whether valid counterevidence produces an appropriate reversal.

These tests establish where each evidence profile is reliable and where its authority ends. Combining profiles requires a separate analysis because they may rest on different claims and assumptions.

\section*{INTEGRATING EVIDENCE ACROSS THE MEDIA LIFECYCLE}

Real cases rarely contain one signal. A detector may assess properties of the payload, a watermark may identify a service, and a signed record may describe one part of the production path. These signals cannot be averaged as though they were repeated measurements of the same fact. Integration begins by preserving which claim each signal supports, where it applies, and what common dependency could make several signals fail together.\cite{c2pa2026spec,golaszewski2026provenance}

\subsection*{Claim-centered case structure}

For an asset, we organize the case as a claim graph with three kinds of nodes: signals contain observations and records, claims state the propositions being assessed, and actions represent possible decisions. Links record support, contradiction, and shared dependence. This graph is a detailed implementation of the short path in Equation~\eqref{eq:evidence-chain}, not a separate scoring model.

The graph prevents support for one proposition from being treated as support for another. A detector can support synthetic origin without identifying a source. A service record can identify where one version was created without establishing permission, and a signed manifest authenticates its assertions rather than the depicted event. The same case can contain supported, contradicted, and unresolved claims.\cite{c2pa2026spec,c2pa2025explainer}

\subsection*{Dependence and localization}

Signals corroborate one another most strongly when they observe different parts of the pipeline. A payload detector and a signed service record may do so; two detectors trained on the same data, or two records issued through the same authority, may not. Simple voting can otherwise count one shortcut or one compromised issuer several times.\cite{yan2025sanity,nemecek2026contradictions}

Signals that share a training source, issuer, or vulnerability can be grouped and assessed jointly. Several agreeing outputs from one dependency group should not count as several independent witnesses. Apparent agreement carries less weight when it can be traced to the same underlying failure mode.

Signals must also remain local to the content they describe. A video may be captured in one interval and generated in another, while its soundtrack follows a third history. Evidence should remain attached to the relevant region, interval, or modality. Localized marks and edit records can preserve that binding; the claim graph records how the components contribute to the finished work.\cite{watermark2024AnythiLocaliMessag,gan2025genptw}

\subsection*{Uncertainty and credential status over time}

Missing evidence does not have one meaning. A signal may never have been issued, may have disappeared during circulation, or may be present but no longer valid. Collapsing these states into ``no credential'' creates false suspicion and an easy route for evasion. A camera that never issued credentials is not equivalent to a marked output whose signal was stripped. Incomplete histories need an explicit unknown state.\cite{c2pa2026spec,golaszewski2026provenance}

Conflict is a separate state. A manifest and watermark may name different services, or the payload may be inconsistent with the declared history. Authenticating one record does not reconcile it with an incompatible assertion. Suppressing the contradiction would turn uncertainty into false confidence.\cite{nemecek2026contradictions,c2pa2026spec}

Time changes the meaning of a credential. A revoked key may indicate later compromise without invalidating every earlier asset, just as a withdrawn license may affect future reuse without rewriting the original creation history. Each evidentiary result needs its own timestamp, method version, calibration record, and credential status. Without that history, a once-valid conclusion can silently outlive the conditions that justified it.\cite{c2pa2026spec}

\subsection*{Proportionate and revisable decisions}

Evidence thresholds should rise with the consequence of the decision. A sensitive signal may be enough to prioritize inspection, but restrictions on reach or monetization require stronger calibration and review. Public attribution demands the strongest corroboration because the cost of a false conclusion is high and difficult to reverse.

Decision records show what the evidence supports, what remains uncertain, and which policy turned that assessment into action. Labels need the same precision because ``AI-generated,'' ``credentials unavailable,'' and ``edited after capture'' describe different states. The interface can expose the bounded claim without forcing every viewer to inspect the full technical record.\cite{gamage2025labeling,feng2023provenance,trattner2026c2palabels}

An affected party also needs a route to add counterevidence. The case record should show how the new information changes the graph and why the decision is retained or reversed. Reproducing the first result is not enough; the system must also be able to correct it.

\subsection*{Worked cases across the lifecycle}

\textit{News and event verification.} A newsroom receives a recompressed viral video with no attached credentials. Frame and temporal detectors identify intervals for closer inspection.\cite{jia2026coda,li2025realworldaigid,li2025aegis} Reporters then trace the source and compare recoverable records with the claimed place and time. A verified capture device can support origin while the accompanying caption remains unresolved. The editorial record keeps those conclusions separate.

\textit{Service attribution in commercial media.} A generated advertisement carries an invisible mark that leads to a signed service record. Matching the two can establish which service produced this version.\cite{fernandez2023stable,gan2025genptw,c2pa2026spec} It does not establish that the advertiser was entitled to publish it. If embedded and remote records disagree, the conflict remains part of the case rather than being resolved by whichever score is larger.\cite{black2025BoxForgerAttack,nemecek2026contradictions}

\textit{Creator or identity dispute.} An artist or individual challenges an advertisement that imitates a protected work or likeness. Earlier records establish what is claimed and when it existed; a technical test examines whether the disputed model or output can be connected to it. Dataset audits become unstable when little protected material is marked or many claimants are tested, so their result cannot carry the case alone.\cite{shao2025databench,ren2026dwbench} The response follows the relationship that can actually be supported and remains open to later evidence.

Across these cases, different tools must preserve the relationship between an asset component, a claim, and its supporting record. Detection remains the fallback when no record survives. A mark can reconnect an asset to a richer history, which may then be compared with authorization records. Institutions still decide what the evidence means for the case, but the technical record can keep the basis and limits of that decision open to inspection.\cite{golaszewski2026provenance,shao2025databench,ren2026dwbench}

\section*{DISCUSSION AND RESEARCH AGENDA}

The comparisons in this review reveal three bottlenecks that are often described as one robustness problem. Passive detectors fail when the observable signal changes. Proactive methods fail when participation, binding, or record access breaks. Rights mechanisms fail when the affected party cannot supply enough protected material or obtain meaningful access to the model. Because these failures occur at different points in the evidence chain, no single robustness measure can capture them.

\subsection*{Synthetic status as a weak endpoint}

As generation becomes an ordinary editing operation, the label ``synthetic'' carries less information on its own. A photographed scene can contain a generated object; a real performance can be paired with synthetic speech; a fully generated advertisement can be authorized and accurately labeled. The useful question is increasingly which operation materially changed the part of the media relevant to the decision.

This shift changes the endpoint of detection research. Binary origin estimates remain valuable when no history survives, but they become an entry point for investigation rather than a complete account of authenticity. Benchmarks can reflect this by distinguishing full synthesis, local generation, semantic editing, and harmless computational assistance, then attaching the result to the affected region or interval. The materiality threshold belongs to the claim, not to the mere presence of an AI operation.\cite{nguyen2026safe,li2025aegis,aigvdbench2025}

\subsection*{Continuity beyond signal robustness}

Robustness asks whether a signal can still be recovered after a transformation. Evidentiary continuity asks a harder question: after that transformation, does the recovered signal still support the same claim about the same component? A watermark copied to another image can be robust but misleading. A manifest retained after an undeclared edit can remain cryptographically valid while no longer describing the visible asset completely.

Benchmarks should evaluate transformation trajectories rather than isolated before-and-after files. A hidden operation log can record compression, cropping, semantic editing, regeneration, insertion into video, and recapture. At each stage, the verifier recovers only the claims that remain justified. A continuity curve would show where a payload binding, remote record, or credential stops supporting its original proposition, instead of compressing that history into one robustness average.\cite{lu2025vine,dixit2026peccavi,c2pa2026spec,golaszewski2026provenance}

Video and interactive media make continuity stateful. The object of verification is no longer one immutable file but a sequence of versions, components, and user actions. Persistent component identifiers, time-bounded assertions, and open mappings between local signals and remote records become more important than a stronger file-level mark.\cite{chen2024demamba,li2025aegis,aigvdbench2025,meta2024audioseal,c2pa2026spec}

\subsection*{Why more signals do not always add trust}

Layered evidence is useful only when the layers fail differently. Two detectors trained on the same benchmark and backbone are not two independent witnesses. A watermark and manifest issued by the same provider can share an account system, key registry, and revocation service. Agreement then reflects common infrastructure as much as independent corroboration.

This dependence creates concentration risk. A widely adopted issuer can make verification easier while also becoming a common point of failure or exclusion. Benchmark reports can make that risk visible by naming shared training sources, foundation encoders, registries, and authorities. Stress tests can then remove or compromise one dependency and measure how much of the conclusion remains supported.\cite{yan2025sanity,nemecek2026contradictions,golaszewski2026provenance}

Independence also has an institutional dimension. A result that only the model provider can reproduce is not equivalent to evidence available to an affected creator, journalist, or court. Access, privacy, and review cost determine who can challenge the record and how much practical authority the signal deserves.

\subsection*{Contestability as a technical property}

Verification is often evaluated at the moment of the first decision. Real disputes continue after that point. New records appear, keys are revoked, a claimant supplies authorization, or an affected party shows that the detector was applied outside its calibrated domain. A trustworthy system retains enough state to explain the first decision and revisit it without erasing the earlier record.

This makes appeal and correction measurable. An evaluation can introduce valid counterevidence and record whether the claim, explanation, and action change appropriately. It can also measure the time and cost required for an independent party to trigger that review. A system that is accurate but practically impossible to challenge concentrates authority rather than establishing trust.\cite{gamage2025labeling,feng2023provenance,trattner2026c2palabels}

The same infrastructure raises a privacy constraint. Persistent attribution can help resolve misuse while exposing creators and ordinary users to tracking. Evidence records need selective disclosure, purpose limits, and retention rules alongside durable verification. The goal is not to expose the entire history to every verifier, but to reveal enough to support the bounded claim at issue.

\subsection*{A measurable research program}

A near-term testbed could follow the same assets through several lifecycles while retaining a hidden operation history. Detection systems, watermark verifiers, provenance readers, and rights audits would each receive the evidence appropriate to their role. Routine distribution, an adaptive attack, a conflicting record, and valid counterevidence would be introduced in separate stages. This design allows each method to answer its own question while exposing shared dependencies.

The resulting metrics would describe more than signal recovery. Claim calibration measures whether confidence matches the stated proposition. Continuity measures how long the asset--claim binding survives. Dependency-adjusted corroboration discounts evidence with a common failure source. Reversal correctness measures whether valid counterevidence changes the outcome, and access cost records who can realistically obtain review. Together, these measures connect technical performance to the settings in which the result informs a consequential decision.

Over the longer term, evidence should be able to outlive one model or platform without becoming a universal tracking layer. This will require interoperable component identifiers, versioned verification procedures, privacy-preserving records, and independent routes for review. Progress should be measured by whether a system preserves a justified claim through change, acknowledges when the connection breaks, and corrects consequential decisions when stronger evidence arrives.

\section*{CONCLUSIONS}

Synthetic media are difficult to verify because the visible asset reveals only part of their history. We have treated each conclusion about that history as a bounded claim supported by observable evidence. Detection remains essential when only the payload survives. Recorded provenance can preserve a stronger connection to production, while rights mechanisms ask what that history means for the people affected by it. These approaches meet at different points in the same lifecycle rather than competing to produce one universal authenticity score.

Across the literature, clean accuracy remains a poor guide to evidentiary value. Detector performance declines when the synthesis process or acquisition channel changes. Watermarks and manifests are stronger only when their records remain reachable and verifiable, and rights audits are constrained by the limited access available to affected people. A result becomes meaningful when its claim and operating conditions are explicit. The framework, tables, and lifecycle protocol developed here provide a common way to describe those conditions.

Future infrastructure must preserve evidence as media changes without extending a narrow result to the entire asset. Each claim should remain tied to the relevant component and moment; dependencies, disagreements, and missing records should remain visible. The same infrastructure must support dynamic media without turning attribution into routine surveillance. Most importantly, consequential decisions must remain open to revision when stronger evidence appears.

\section*{DECLARATION OF INTERESTS}

The authors declare no competing interests.

\begingroup
\fontsize{10.3}{11.5}\selectfont
\setlength{\bibsep}{0pt}
\sloppy
\bibliography{references_300plus}
\endgroup

\end{document}